\documentclass[runningheads]{llncs}
\usepackage{enumitem}
\usepackage[T1]{fontenc}
\usepackage{multirow}
\usepackage{amssymb}
\usepackage{float}
\usepackage{graphicx}

\begin{document}
\emergencystretch=2em

\title{Location-Aware Language Models via Secondary Embeddings}

\author{Gokul Srinivasagan \and
Munir Georges}

\institute{AImotion Bavaria, Technische Hochschule Ingolstadt, Germany \\
\email{\{Gokul.Srinivasagan, Munir.Georges\}@thi.de}\\
}

\maketitle             

\begin{abstract}
Pretrained transformer-based language models achieve strong performance across a wide range of NLP tasks but remain limited in encoding geo-locational semantics, leading to suboptimal representations of place names and spatial entities. In this work, we propose a lightweight, model-agnostic approach for injecting geo-spatial awareness into pretrained embeddings without modifying the tokenizer or requiring costly retraining. Our method augments input representations with structured geographic signals by combining location names with their corresponding latitude and longitude, and employs a location-focused masking to better align textual representations with real-world spatial relationships. This design allows the model to incorporate geo-spatial context while preserving existing semantic and syntactic knowledge. 
Experimental results demonstrate substantial improvements in geo-spatial alignment while maintaining comparable performance on standard NLP benchmarks such as GLUE. The method is computationally efficient, requiring only minutes of additional training, and generalizes across multiple model architectures and scales.
\end{abstract}

\section{Introduction}

Recent advances in pretrained language models such as BERT \cite{devlin2019bert} and RoBERTa \cite{liu2019roberta} have demonstrated a remarkable ability to capture semantic and syntactic relationships between words. Despite this success, these models often struggle with specialized domains that require additional contextual grounding. In particular, geo-locational awareness is essential for understanding spatial relationships and providing contextually relevant outputs such as route descriptions, targeted advertisements, and localized search results. However, existing models largely fail to encode geographic semantics effectively for entities such as cities, landmarks, or regions.

\begin{figure}[!ht]
    \centering
    \includegraphics[width=.85\textwidth]{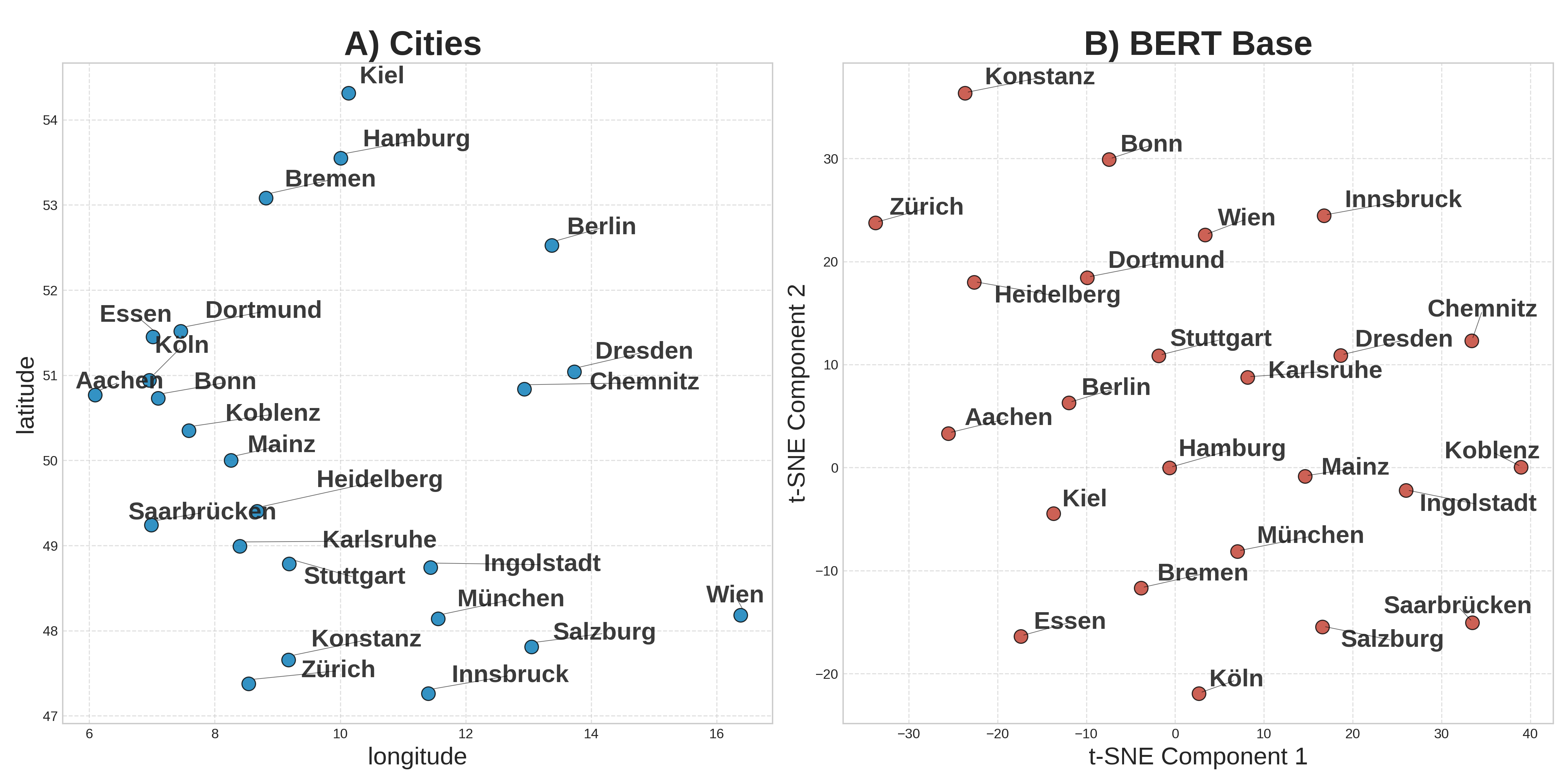}
    \caption{Comparison between the actual geographic location of cities (left) and their corresponding t-SNE visualization in the BERT embedding space (right). 
    The visualization shows that cities which are geographically distant often appear close together in the embedding space, highlighting the lack of geo-spatial awareness in pretrained language models.}
    \label{fig:loc_emb}
\end{figure}

As illustrated in Figure~\ref{fig:loc_emb}, embeddings of city names are not well structured in the latent space, largely due to the absence of explicit geographic grounding in language models. For instance, the cities \textit{Saarbr\"ucken} and \textit{Salzburg} appear close to each other in BERT’s embedding space, even though they are geographically distant. This discrepancy highlights the model’s inability to reflect real-world spatial relationships.

A contributing factor to this limitation lies in the tokenization process. While subword tokenization has proven effective for general text representation, it fragments named entities, eroding their unique semantic identity. For example, the city name \textit{Ingolstadt} is split into subwords \textit{"ing"}, \textit{"\#\#ols"}, and \textit{"\#\#tadt"}, resulting in partial embeddings that fail to capture the entity as a single coherent concept. The issue worsens when different locations share identical names (e.g., Paris, France vs.\ Paris, Texas), as the model has no geo-spatial information to distinguish between them. Consequently, embeddings of distinct entities often collapse into the same region of the vector space, misrepresenting their relational proximity.

Existing approaches to incorporating geographic information often require architectural modifications or costly retraining procedures, limiting their scalability and practical applicability \cite{li-etal-2022-spabert,li-etal-2023-geolm,ding2023mgeo}. A seemingly straightforward solution would be to expand the tokenizer’s vocabulary to include all place names. However, this approach is impractical as it significantly increases model size and training costs. For instance, adding merely 1,000 new tokens to the BERT-base vocabulary increases the embedding parameters by approximately 3.2\%. Moreover, retraining such a model to integrate the new tokens into its representational space is computationally expensive and may degrade performance on downstream NLP tasks.

To address these limitations, we propose a lightweight and model-agnostic method for integrating geographic information into pretrained embeddings without modifying the tokenizer or requiring large-scale retraining. Our approach injects structured geo-spatial signals into the embedding space, aligning representations with real-world spatial relationships while preserving existing linguistic knowledge. As a result, the model gains location awareness with minimal computational overhead and no loss in downstream performance.

The main contributions of this work are as follows:
\begin{itemize}[noitemsep,topsep=0pt]
    \item We propose a lightweight method for integrating geo-locational information into pretrained language model embeddings without modifying the tokenizer or vocabulary.
    \item Our approach enriches the embedding space with structured geographic signals while preserving existing syntactic and semantic knowledge.
    \item We demonstrate substantial improvements in geo-spatial awareness without degrading performance on downstream benchmarks such as GLUE.
    \item We also validate the generalizability of our method across different model architectures, scales, and multilingual settings.
\end{itemize}

\section{Related Work}

Prior work has shown that incorporating geographic information into language models can improve the representation of spatial entities and performance on geo-text understanding tasks \cite{li-etal-2022-spabert,li-etal-2023-geolm,ding2023mgeo,TALE}. SpaBERT \cite{li-etal-2022-spabert} employs a masked entity prediction objective alongside masked language modeling to capture geographic relationships among entities. GeoLM \cite{li-etal-2023-geolm} combines masked language modeling with contrastive learning to jointly model linguistic and geographic information. Similarly, MGeo \cite{ding2023mgeo} introduces a multimodal architecture with a dedicated geographic encoder for query and point-of-interest (POI) matching. While these approaches demonstrate the benefits of geo-spatial modeling, they depend on specialized architectures, additional learning objectives, or extensive training to acquire location awareness.

In contrast, our method injects geographic signals into pretrained language models through lightweight input augmentation, requiring no tokenizer modifications, no additional learning objectives beyond masked language modeling, and only a short period of training, while preserving general semantic knowledge.

\section{Method}

\begin{figure*}[!ht]
    \centering
    \includegraphics[width=.95\textwidth]{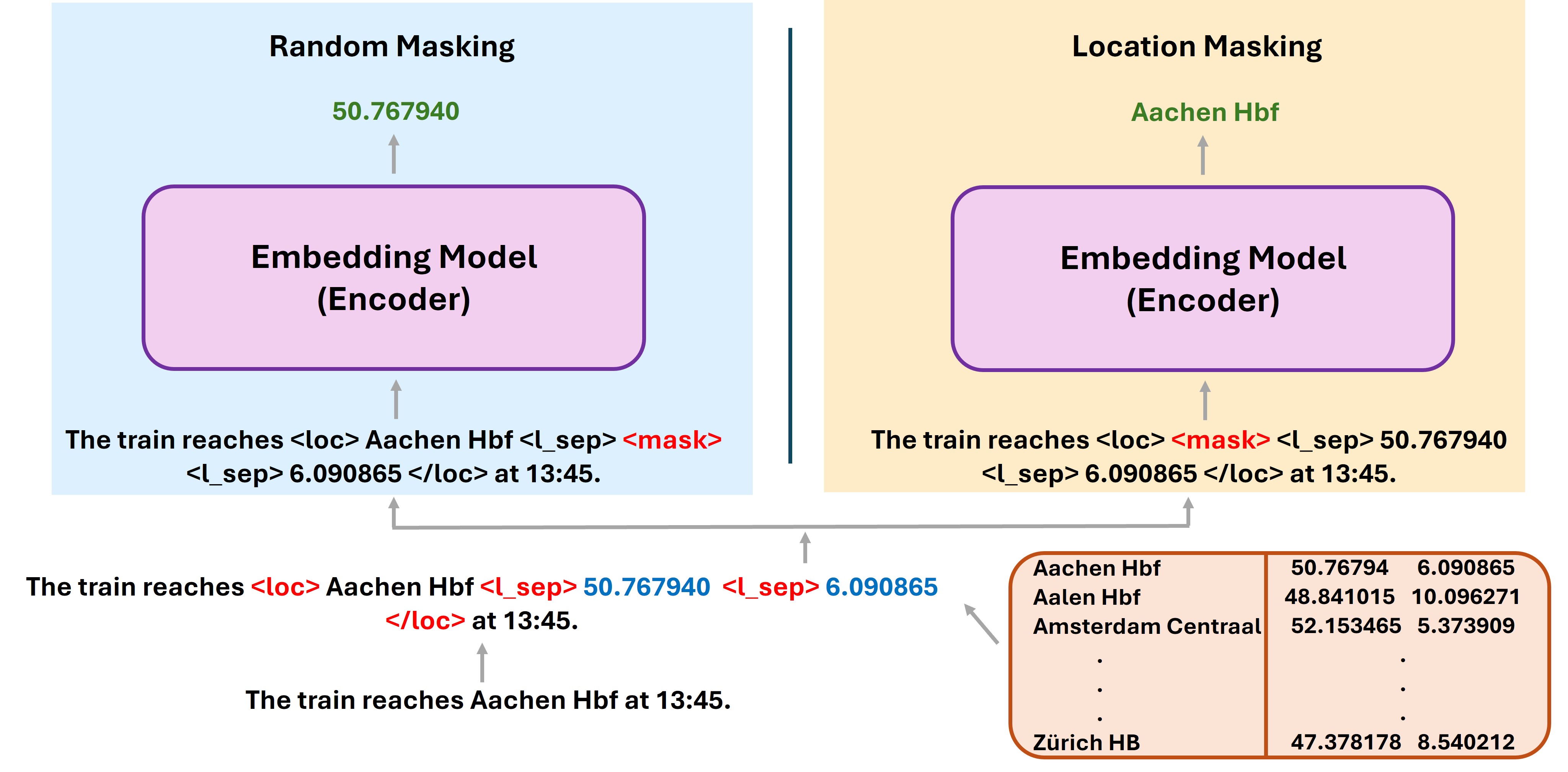}
    \caption{Overview of the proposed framework for incorporating geo-spatial information into pretrained language model embeddings. 
    Given an input sentence containing location entities, we construct a secondary embedding that encodes the corresponding geographic information (city name, latitude, and longitude). 
    This information is integrated into the input sequence using special tokens (\textit{$<loc>$}, \textit{$</loc>$}, and \textit{$<l\_sep>$}) to inject structured location signals into the model. 
    During training, we employ two masking strategies: (1) standard masked language modeling (MLM) with random token masking (left), and (2) a location-focused masking scheme that selectively masks city names (right). 
    The model is trained to recover masked entities using both contextual and geo-spatial cues, enabling improved location-aware representations.}
    \label{fig:loc_emb_arch}
\end{figure*}

\subsection{Dataset Generation}
\label{sec:dataset}

We construct a synthetic dataset using information derived from the General Transit Feed Specification (GTFS) for long-distance train schedules in Germany\footnote{https://gtfs.de/en/feeds/}. After pre-processing, we extract 820 unique location names corresponding to train stations, which serve as geographic entities. 

To generate natural language training data, we employ two large language models: Mistral \cite{Jiang2023Mistral7,jiang2024mixtral} and LLaMA 3 \cite{grattafiori2024llama}, to produce journey descriptions containing station names, arrival and departure times, and other contextual details. This results in a corpus of 5,965 text samples with a total of 62,758 occurrences of location entities.

In addition to textual data, we associate each location with its corresponding latitude and longitude coordinates, forming a structured geo-spatial reference that is later incorporated into the training process.

\subsection{Continual Pretraining with Secondary Embeddings}

To enable geo-spatial awareness, we use three special tokens: 
\textit{$<loc>$} (start of location), \textit{$</loc>$} (end of location), and \textit{$<l\_sep>$} (location separator). These tokens explicitly define the boundaries of location-specific information within the input sequence.

We construct a mapping between each \textit{stop\_name} and its geographic coordinates (latitude and longitude), which we refer to as \textit{secondary embeddings}. Unlike standard token embeddings learned purely from text, these embeddings encode structured spatial information and act as an auxiliary signal during training.

During tokenization, each occurrence of a location entity is augmented into the following structured format:

\begin{center}
\textbf{$<loc>$} \textit{stop\_name} \textbf{$<l\_sep>$} \textit{latitude} \textbf{$<l\_sep>$} \textit{longitude} \textbf{$</loc>$}
\end{center}

This representation injects both textual and coordinate-based information directly into the model’s input stream, enabling the model to associate entities with their corresponding spatial attributes.

We perform continual pretraining using the masked language modeling (MLM) objective with two different masking strategies (Figure \ref{fig:loc_emb_arch}):

\begin{itemize}
    \item \textbf{Random Masking ($\mathcal{R}$):} Following the standard BERT training procedure \cite{devlin2019bert}, a subset of tokens is randomly masked.
    
    \item \textbf{Location Masking ($\mathbb{L}$):} We specifically mask the \textit{stop\_name} tokens within the \textit{$<loc>$} blocks while retaining the associated latitude and longitude information. The model is trained to predict the masked location entity using both contextual and geo-spatial cues. This targeted objective encourages alignment between textual representations and their underlying geographic structure.
\end{itemize}

We train separate models under each masking strategy and compare their effectiveness in inducing geo-spatial awareness.

\section{Experiment Details}

\subsection{Evaluation}
The primary objective of our approach is to enrich pretrained language model embeddings with geo-locational information while preserving their existing semantic knowledge. To verify that location-aware pretraining does not degrade linguistic capability, we evaluate the models on the GLUE benchmark \cite{wang-etal-2018-glue}. We report F1-scores for MRPC, Pearson correlation for STS-B, and accuracy for the remaining tasks. Performance comparable to the baseline indicates that the injected geo-spatial information does not compromise general language understanding.

To evaluate location awareness, we analyze the alignment between distances in the embedding space and real-world geographic distances. We compute pairwise geographic distances between locations using the Haversine formula based on latitude and longitude coordinates, and treat this as the ground-truth distance matrix. 

Since model embeddings reside in a high-dimensional space, we measure distances between location embeddings using Euclidean distance and cosine similarity. Our empirical evaluation on ground-truth data demonstrates a strong correlation between embedding-based distances and Haversine distance, indicating that Euclidean distance and cosine similarities provide reliable proxies for geographic distance in the embedding space.

We then compute Pearson and Spearman correlation coefficients between the geographic distance matrix and the corresponding embedding distance matrices. Higher correlation indicates better preservation of real-world spatial relationships in the learned embedding space.

\subsection{Implementation Details}

We train the models on the synthetic dataset described in Section \ref{sec:dataset} for 25 epochs using a batch size of 60 and a learning rate of $1\times10^{-4}$ with the AdamW optimizer. We use pretrained models using the Hugging Face Transformers library \cite{wolf-etal-2020-transformers}.

For GLUE tasks, fine-tuning is performed for up to 50 epochs with a batch size of 256 and a learning rate of $1\times10^{-5}$ using bfloat16 precision. We apply early stopping with a patience of 5 to prevent overfitting.

Unless otherwise specified, all experiments are conducted using the base variant of each pretrained model. Location-aware training completes within approximately 15 minutes on a single NVIDIA H100 GPU, demonstrating the computational efficiency of the proposed approach.

\begin{table*}[!ht]
    \centering
    \caption{GLUE benchmark results for baseline and location-aware models across multiple architectures. Here, $\mathcal{R}$ and $\mathbb{L}$ denote models trained with random masking and location-specific masking, respectively. The results show that location-aware training preserves general language performance while achieving comparable or slightly improved scores.}
    \begin{tabular}{@{}l@{}|lllllllll@{}|l@{}} 
    \hline
        \textbf{Model} & \textbf{cola} & \textbf{mrpc} & \textbf{qnli} & \textbf{qqp} & \textbf{rte} & \textbf{sst2} & \textbf{stsb} & \textbf{wnli} & \textbf{mnli} & \textbf{Total} \\ \hline

        \textbf{BERT$_{base}$} & 81.78 & 86.00 & 90.74 & 90.73 & 58.84 & 93.00 & 88.94 & 56.34 & 84.04 & 81.16  \\
        \textbf{{BERT$_{base}$}($\mathcal{R}$)} & 82.07 & 86.39 & 90.99 & 90.27 & 64.26 & 91.28 & 89.68 & 56.34 & 84.13 & 81.71  \\ 
        \textbf{{BERT$_{base}$}($\mathbb{L}$)} & 81.78 & 87.58 & 91.12 & 90.28 & 63.89 & 92.20 & 89.87 & 56.34 & 84.55 & 81.96  \\ \hline

        \textbf{DistilBERT} & 79.39 & 89.13 & 88.16 & 89.42 & 60.29 & 90.48 & 87.03 & 56.34 & 81.31 & 80.17  \\ 
        \textbf{DistilBERT($\mathcal{R}$)} & 79.58 & 86.12 & 88.94 & 89.60 & 59.93 & 90.60 & 86.80 & 54.93 & 81.47 & 79.77  \\ 
        \textbf{DistilBERT($\mathbb{L}$)}& 78.23 & 87.92 & 87.68 & 89.66 & 59.93 & 90.37 & 86.99 & 56.34 & 81.42 & 79.84  \\ \hline
        
        \textbf{RoBERTa} & 81.87 & 91.80 & 92.11 & 90.33 & 75.89 & 92.12 & 90.19 & 56.34 & 85.99 & 84.07  \\ 
        \textbf{RoBERTa($\mathcal{R}$)} & 81.98 & 91.74 & 92.11 & 90.43 & 73.65 & 92.66 & 90.47 & 56.34 & 85.96 & 83.93 \\ 
        \textbf{RoBERTa($\mathbb{L}$)} & 83.22 & 91.41 & 91.85 & 90.52 & 71.12 & 93.35 & 90.52 & 56.34 & 86.85 & 83.91 \\ \hline

        \textbf{XLM-R} & 69.13 & 90.58 & 89.97 & 90.28 & 63.90 & 92.32 & 88.15 & 56.34 & 83.35 & 80.45  \\ 
        \textbf{XLM-R($\mathcal{R}$)} & 77.37 & 88.44 & 90.04 & 90.71 & 64.62 & 91.63 & 85.79 & 56.34 & 83.19 & 80.90  \\ 
        \textbf{XLM-R($\mathbb{L}$)} & 78.14 & 88.97 & 89.68 & 90.52 & 52.71 & 91.06 & 85.10 & 56.34 & 82.81 & 79.48  \\ \hline

        \textbf{mBERT} & 75.07 & 88.97& 90.54 & 89.50 & 65.70 & 90.37 & 88.28 & 56.34 & 81.30 & 80.67  \\ 
        \textbf{mBERT($\mathcal{R}$)} & 75.17 & 88.73 & 90.48 & 89.51 & 67.51 & 88.65 & 88.12 & 56.34 & 81.01 & 80.61  \\ 
        \textbf{mBERT($\mathbb{L}$)} & 75.36 & 87.15 & 90.68 & 89.42 & 67.51 & 88.88 & 87.47 & 54.93 & 81.20 & 80.29  \\ \hline
       
        \textbf{ALBERT} & 69.13 & 89.86 & 89.91 & 90.07 & 70.40 & 91.97 & 91.06 & 56.34 & 84.20 & 81.44  \\ 
        \textbf{ALBERT($\mathcal{R}$)} & 66.63 & 77.78 & 90.48 & 89.94 & 52.71 & 90.02 & 89.64 & 56.34 & 83.59 & 77.46  \\ 
        \textbf{ALBERT($\mathbb{L}$)} & 74.40 & 87.94 & 90.87 & 90.13 & 72.56 & 91.40 & 90.87 & 56.34 & 84.13 & 82.07  \\ \hline
        
    \end{tabular}
    \label{tab:glue_results}
\end{table*}

\section{Results and Discussion}

Table~\ref{tab:glue_results} presents the GLUE benchmark results for our location-aware models across multiple architectures. Overall, the results show that the approach preserves general language understanding, with performance comparable to baseline pretrained models across all tasks.

A closer inspection of Table~\ref{tab:glue_results} shows that location-aware variants often match or slightly improve upon baseline performance. For instance, BERT$_{base}$ improves from 81.16 to 81.96 with location masking, while ALBERT shows a noticeable gain from 81.44 to 82.07. These improvements indicate that injecting geo-spatial information does not interfere with linguistic knowledge and, in some cases, provides additional useful signals.

Across architectures, location-specific masking ($\mathbb{L}$) tends to yield more consistent gains compared to random masking ($\mathcal{R}$), particularly for BERT \cite{devlin2019bert} and ALBERT \cite{Lan2020ALBERT}. This suggests that explicitly masking location entities creates a stronger and more targeted learning signal for aligning textual and geo-spatial representations. In contrast, random masking primarily preserves baseline performance without consistently improving it.

\paragraph{Embedding Analysis.}
To evaluate how location information is encoded, we compare two embedding extraction strategies: the \textit{[CLS]} token and mean pooling over all tokens. While both capture location-aware signals, mean pooling consistently exhibits stronger alignment with geographic structure. For example, in RoBERTa, the Pearson correlation improves from 0.2997 (baseline) to 0.3343 using \textit{[CLS]}, and further to 0.4503 using mean embeddings. This indicates that geo-spatial information is distributed across tokens rather than concentrated in a single representation.

\paragraph{Impact of Tokenization.}
We observe that tokenizer choice significantly influences the effectiveness of location-aware adaptation. Models based on WordPiece tokenization (e.g., BERT and DistilBERT) show more stable and consistent improvements compared to byte-level BPE (RoBERTa) and SentencePiece-based models (XLM-R, mBERT). WordPiece decomposes rare words into meaningful subword units, which helps preserve partial semantic structure of location names. In contrast, byte-level BPE operates at the character level, and SentencePiece treats text as a continuous Unicode stream, which may dilute the representation of structured entities such as location names and numeric coordinates.

Interestingly, uncased models tend to benefit more consistently from location-aware training, whereas cased models show less pronounced improvements. This suggests that case sensitivity may introduce additional variability in entity representations, making it harder to learn consistent geo-spatial relationships.

\subsection{Effect of Model Size}

\begin{table}[!ht]
     \caption{Correlation results comparing embedding distances with geographic distances. $r$ and $\rho$ denote Pearson and Spearman correlation coefficients, respectively. Higher values indicate improved geo-spatial alignment.}
    \centering
    \begin{tabular}{@{}l@{}|c|c|c|c@{}}
    \hline
        \multirow{2}{*}{\textbf{Model}} & \multicolumn{2}{c|}{\textbf{Euclidean}} &  \multicolumn{2}{c}{\textbf{Cosine sim}}  \\ \cline{2-5}
         & \textbf{$r$} & \textbf{$\rho$} & \textbf{$r$ }& \textbf{$\rho$}  \\ \hline
         
        \textbf{BERT$_{base}$ }& 0.316 & 0.300 & 0.309 & 0.307  \\ 
        \textbf{{BERT$_{base}$}($\mathcal{R}$)} & 0.371 & 0.345 & 0.336 & 0.316  \\ 
        \textbf{{BERT$_{base}$}($\mathbb{L}$)} & \textbf{0.450} & \textbf{0.431} & \textbf{0.390} & \textbf{0.368}  \\ \hline
        \textbf{{BERT$_{large}$}} & 0.297 & 0.326 & 0.273 & 0.287  \\ 
        \textbf{{BERT$_{large}$}($\mathcal{R}$)} & 0.360 & 0.345 & 0.314 & 0.288  \\ 
        \textbf{{BERT$_{large}$}($\mathbb{L}$)} & \textbf{0.550} & \textbf{0.518} & \textbf{0.422} & \textbf{0.396}  \\ \hline
        \textbf{{DistilBERT}} & 0.375 & 0.345 & 0.348 & 0.329  \\ 
        \textbf{{DistilBERT}($\mathcal{R}$)}& 0.420 & 0.394 & 0.356 & 0.330  \\ 
        \textbf{{DistilBERT}($\mathbb{L}$)} & \textbf{0.453} & \textbf{0.413} & \textbf{0.423} & \textbf{0.377}  \\ \hline
    \end{tabular}
    \label{tab:loc_results}
\end{table}

To study the impact of model scale, we evaluate BERT variants of different sizes, including BERT$_{base}$, BERT$_{large}$, and DistilBERT. The configurations of these models are shown in Table \ref{tab:models}.

\begin{table}[H]
    \caption{Configurations of the models used in our experiment.}
    \centering
    \begin{tabular}{@{}l@{}|llll@{}}
     \hline
        \textbf{Model}   & \textbf{\#Layer} & \textbf{Dim.} & \textbf{\#Attn} & \textbf{\#Param}\\ \hline
        \textbf{BERT$_{base}$}   & 12    & 768  & 12 &  110M \\
        \textbf{BERT$_{large}$}   & 24     & 1024  & 16 & 336M \\
        \textbf{DistilBERT} &   6     & 768  & 12 & 66M \\
        \hline       
    \end{tabular}
    \label{tab:models}
\end{table}

Table~\ref{tab:loc_results} reports the correlation between embedding distances and ground-truth geographic distances using both Euclidean and cosine similarity. The results show that larger models benefit more from location-aware training. BERT$_{large}$ achieves the highest correlation (Pearson $r = 0.550$), indicating a stronger alignment between embedding space and real-world geography. BERT$_{base}$ and DistilBERT also show substantial improvements over their respective baselines, demonstrating that the proposed method generalizes across model scales.

We further evaluate a smaller model, BERT-tiny \cite{turc2019well}, which shows only marginal improvement (from 0.0123 to 0.0139). This suggests that very small models lack sufficient capacity to effectively incorporate geo-spatial signals.

Additional experiments with RoBERTa \cite{liu2019roberta} show a significant improvement in correlation (from 0.2665 to 0.4296), further confirming the effectiveness of our approach in larger architectures.

\subsection{Effect of Multilingual Models}

We evaluate the impact of location-aware training on multilingual models, including mBERT \cite{devlin-etal-2019-bert} and XLM-R \cite{conneau-etal-2020-unsupervised}. In contrast to monolingual models, we observe limited or no consistent improvement in geo-spatial alignment.

This can be attributed to the significantly larger and more diverse vocabularies used in multilingual models (e.g., $\sim$105k tokens for mBERT and $\sim$250k for XLM-R), as well as the increased variability in entity representation across languages. These factors make it more challenging for the model to learn consistent mappings between textual and geographic representations.

\subsection{Effect of Other Architectures}

We further evaluate the generalizability  of our approach on alternative architectures such as ALBERT \cite{Lan2020ALBERT} and MobileBERT \cite{sun-etal-2020-mobilebert}. For ALBERT, we observe a substantial improvement in geo-spatial alignment, with correlation increasing from 0.0039 to 0.1746. Similarly, MobileBERT shows a improvement from 0.0123 to 0.0345. These results demonstrate that our method is not limited to standard transformer architectures and can generalize to parameter-efficient and compressed models.

\section{Conclusion}

In this work, we introduced a lightweight and effective approach for incorporating geo-spatial information into pretrained language models. By augmenting the input with structured location signals and applying targeted masking strategies, our method enables conventional language models to become location-aware without requiring architectural modifications, vocabulary expansion, or costly retraining. Experimental results demonstrate that our approach improves the alignment between the embedding space and real-world geographic relationships, while maintaining comparable performance on standard NLP benchmarks such as GLUE. These findings confirm that location awareness can be introduced without degrading the model’s existing semantic and syntactic capabilities.
Furthermore, the proposed method is computationally efficient, requiring only around 15 minutes of training to adapt pretrained models. This makes it a practical and scalable solution for enhancing language models with geo-spatial grounding in real-world applications.

\section{Limitations}

Despite the promising results, several limitations remain. First, while our approach improves geo-spatial alignment, a deeper understanding of how location information is encoded within the embedding space—particularly in interaction with semantic and syntactic features—requires further investigation.
Second, our method operates as a post-hoc adaptation to pretrained models. Since most representational capacity is learned during the initial pretraining phase, incorporating geo-spatial information directly during large-scale pretraining may yield stronger and more robust representations. Investigating this direction constitutes an important area for future research.
Finally, our current work focuses on encoder-based models. Extending this approach to generative language models remains an open challenge and an important direction for future exploration.

\section*{Acknowledgments}

This work has been supported by the Verkehrsverbund Großraum Ingolstadt (VGI) as part of the project newMIND.

\bibliographystyle{splncs04}
\bibliography{mybibliography}

@inproceedings{devlin2019bert,
  title={Bert: Pre-training of deep bidirectional transformers for language understanding},
  author={Devlin, Jacob and Chang, Ming-Wei and Lee, Kenton and Toutanova, Kristina},
  booktitle={Proceedings of the 2019 conference of the North American chapter of the association for computational linguistics: human language technologies, volume 1 (long and short papers)},
  pages={4171--4186},
  year={2019}
}

@inproceedings{devlin-etal-2019-bert,
    title = "{BERT}: Pre-training of Deep Bidirectional Transformers for Language Understanding",
    author = "Devlin, Jacob  and
      Chang, Ming-Wei  and
      Lee, Kenton  and
      Toutanova, Kristina",
    booktitle = "Proceedings of the 2019 Conference of the North {A}merican Chapter of the Association for Computational Linguistics: Human Language Technologies, Volume 1",
    year = "2019",
    address = "Minneapolis, Minnesota",
    publisher = "ACL",
    doi = "10.18653/v1/N19-1423",
}

@article{liu2019roberta,
  title={Roberta: A robustly optimized bert pretraining approach},
  author={Liu, Yinhan},
  journal={arXiv preprint arXiv:1907.11692               }     ,
  volume={364},
  year={2019}
}

@inproceedings{sun-etal-2020-mobilebert,
    title = "{M}obile{BERT}: a Compact Task-Agnostic {BERT} for Resource-Limited Devices",
    author = "Sun, Zhiqing  and
      Yu, Hongkun  and
      Song, Xiaodan  and
      Liu, Renjie  and
      Yang, Yiming  and
      Zhou, Denny",
    booktitle = "Proceedings of the 58th Annual Meeting of the Association for Computational Linguistics",
    year = "2020",
    address = "Online",
    publisher = "ACL",
    doi = "10.18653/v1/2020.acl-main.195"
}

@article{Jiang2023Mistral7,
  title={Mistral 7B},
  author={Albert Qiaochu Jiang and Alexandre Sablayrolles and Arthur Mensch and Chris Bamford and Devendra Singh Chaplot and Diego de Las Casas and Florian Bressand and Gianna Lengyel and Guillaume Lample and Lucile Saulnier and L{\'e}lio Renard Lavaud and Marie-Anne Lachaux and Pierre Stock and Teven Le Scao and Thibaut Lavril and Thomas Wang and Timoth{\'e}e Lacroix and William El Sayed},
  journal={ArXiv},
  year={2023},
  volume={abs/2310.06825},
  url={https://api.semanticscholar.org/CorpusID:263830494}
}

@article{jiang2024mixtral,
  title={Mixtral of experts},
  author={Jiang, Albert Q and Sablayrolles, Alexandre and Roux, Antoine and Mensch, Arthur and Savary, Blanche and Bamford, Chris and Chaplot, Devendra Singh and Casas, Diego de las and Hanna, Emma Bou and Bressand, Florian and others},
  journal={arXiv preprint arXiv:2401.04088         },
  year={2024}
}

@article{grattafiori2024llama,
  title={The llama 3 herd of models},
  author={Grattafiori, Aaron and Dubey, Abhimanyu and Jauhri, Abhinav and Pandey, Abhinav and Kadian, Abhishek and Al-Dahle, Ahmad and Letman, Aiesha and Mathur, Akhil and Schelten, Alan and Vaughan, Alex and others},
  journal={arXiv preprint arXiv:2407.21783            },
  year={2024}
}

@article{turc2019well,
  title={Well-read students learn better: On the importance of pre-training compact models},
  author={Turc, Iulia and Chang, Ming-Wei and Lee, Kenton and Toutanova, Kristina},
  journal={arXiv preprint arXiv:1908.08962     }      ,
  year={2019}
}

@inproceedings{
Lan2020ALBERT,
title={ALBERT: A Lite BERT for Self-supervised Learning of Language Representations},
author={Zhenzhong Lan and Mingda Chen and Sebastian Goodman and Kevin Gimpel and Piyush Sharma and Radu Soricut},
booktitle={International Conference on Learning Representations},
year={2020},
url={https://openreview.net/forum?id=H1eA7AEtvS}
}

@inproceedings{conneau-etal-2020-unsupervised,
    title = "Unsupervised Cross-lingual Representation Learning at Scale",
    author = "Conneau, Alexis  and
      Khandelwal, Kartikay  and
      Goyal, Naman  and
      Chaudhary, Vishrav  and
      Wenzek, Guillaume  and
      Guzm{\'a}n, Francisco  and
      Grave, Edouard  and
      Ott, Myle  and
      Zettlemoyer, Luke  and
      Stoyanov, Veselin",
    editor = "Jurafsky, Dan  and
      Chai, Joyce  and
      Schluter, Natalie  and
      Tetreault, Joel",
    booktitle = "Proceedings of the 58th Annual Meeting of the Association for Computational Linguistics",
    month = jul,
    year = "2020",
    address = "Online",
    publisher = "Association for Computational Linguistics",
    url = "https://aclanthology.org/2020.acl-main.747/",
    doi = "10.18653/v1/2020.acl-main.747",
    pages = "8440--8451",
}

@inproceedings{wang-etal-2018-glue,
    title = "{GLUE}: A Multi-Task Benchmark and Analysis Platform for Natural Language Understanding",
    author = "Wang, Alex  and
      Singh, Amanpreet  and
      Michael, Julian  and
      Hill, Felix  and
      Levy, Omer  and
      Bowman, Samuel",
    booktitle = "Proceedings of the 2018 {EMNLP} Workshop {B}lackbox{NLP}: Analyzing and Interpreting Neural Networks for {NLP}",
    year = "2018",
    address = "Brussels, Belgium",
    publisher = "ACL",
    doi = "10.18653/v1/W18-5446"
}

@inproceedings{wolf-etal-2020-transformers,
    title = "Transformers: State-of-the-Art Natural Language Processing",
    author = "Thomas Wolf and Lysandre Debut and Victor Sanh and Julien Chaumond and Clement Delangue and Anthony Moi and Pierric Cistac and Tim Rault and Rémi Louf and Morgan Funtowicz and Joe Davison and Sam Shleifer and Patrick von Platen and Clara Ma and Yacine Jernite and Julien Plu and Canwen Xu and Teven Le Scao and Sylvain Gugger and Mariama Drame and Quentin Lhoest and Alexander M. Rush",
    booktitle = "Proceedings of the Conference on Empirical Methods in NLP: System Demonstrations",
    year = "2020",
    publisher = "ACL"
}

@inproceedings{li-etal-2022-spabert,
    title = "{S}pa{BERT}: A Pretrained Language Model from Geographic Data for Geo-Entity Representation",
    author = "Li, Zekun  and
      Kim, Jina  and
      Chiang, Yao-Yi  and
      Chen, Muhao",
    editor = "Goldberg, Yoav  and
      Kozareva, Zornitsa  and
      Zhang, Yue",
    booktitle = "Findings of the Association for Computational Linguistics: EMNLP 2022",
    month = dec,
    year = "2022",
    address = "Abu Dhabi, United Arab Emirates",
    publisher = "Association for Computational Linguistics",
    url = "https://aclanthology.org/2022.findings-emnlp.200/",
    doi = "10.18653/v1/2022.findings-emnlp.200",
    pages = "2757--2769"
}

@inproceedings{li-etal-2023-geolm,
    title = "{G}eo{LM}: Empowering Language Models for Geospatially Grounded Language Understanding",
    author = "Li, Zekun  and
      Zhou, Wenxuan  and
      Chiang, Yao-Yi  and
      Chen, Muhao",
    editor = "Bouamor, Houda  and
      Pino, Juan  and
      Bali, Kalika",
    booktitle = "Proceedings of the 2023 Conference on Empirical Methods in Natural Language Processing",
    month = dec,
    year = "2023",
    address = "Singapore",
    publisher = "Association for Computational Linguistics",
    url = "https://aclanthology.org/2023.emnlp-main.317/",
    doi = "10.18653/v1/2023.emnlp-main.317",
    pages = "5227--5240"
}

@inproceedings{ding2023mgeo,
  title={Mgeo: Multi-modal geographic language model pre-training},
  author={Ding, Ruixue and Chen, Boli and Xie, Pengjun and Huang, Fei and Li, Xin and Zhang, Qiang and Xu, Yao},
  booktitle={Proceedings of the 46th international ACM SIGIR conference on research and development in information retrieval},
  pages={185--194},
  year={2023}
}

@ARTICLE{TALE,
  author={Wan, Huaiyu and Lin, Yan and Guo, Shengnan and Lin, Youfang},
  journal={IEEE Transactions on Knowledge and Data Engineering}, 
  title={Pre-Training Time-Aware Location Embeddings from Spatial-Temporal Trajectories}, 
  year={2022},
  volume={34},
  number={11},
  pages={5510-5523},
  doi={10.1109/TKDE.2021.3057875}}
\end{document}